\documentclass[conference]{IEEEtran}
\IEEEoverridecommandlockouts
\usepackage{cite}
\usepackage{amsmath,amssymb,amsfonts}
\usepackage{algorithmic}
\usepackage{graphicx}
\usepackage{textcomp}
\usepackage{xcolor}
\usepackage{soul}
\usepackage{nth}
\usepackage{subcaption}
\usepackage{booktabs}
\usepackage{siunitx}
\usepackage{latexsym}
\usepackage{hyperref}
\hypersetup{
    colorlinks=true,
    linkcolor=blue,
    filecolor=magenta,      
    urlcolor=cyan,
}
\usepackage{cleveref}
\crefname{section}{Sec.}{Sections}
\crefname{subsection}{Sec.}{Sections}
\crefname{figure}{Fig.}{Figs.}
\crefname{table}{Tab.}{Tabs.}
\crefname{equation}{Eq.}{Eqs.}
\crefname{appsec}{Appendix}{Appendices}

\makeatletter
    \newcommand{\linebreakand}{%
      \end{@IEEEauthorhalign}
      \hfill\mbox{}\par
      \mbox{}\hfill\begin{@IEEEauthorhalign}
    }
\makeatother
    
\def\BibTeX{{\rm B\kern-.05em{\sc i\kern-.025em b}\kern-.08em
    T\kern-.1667em\lower.7ex\hbox{E}\kern-.125emX}}
\begin{document}

\title{ARIN: Adaptive Resampling and Instance Normalization for Robust Blind Inpainting of Dunhuang Cave Paintings
}

\author{
\IEEEauthorblockN{Alexander Schmidt, %
Prathmesh Madhu\IEEEauthorrefmark{1}, 
Andreas Maier, %
Vincent Christlein and %
Ronak Kosti}%
\IEEEauthorblockA{Pattern Recognition Lab\\
Friedrich-Alexander-Universit\"at Erlangen-N\"urnberg, Germany\\
Email: prathmesh.madhu@fau.edu}
\IEEEauthorblockA{\IEEEauthorrefmark{1}Corresponding Author
}
}

\IEEEoverridecommandlockouts
\IEEEpubid{\makebox[\columnwidth]{978-1-6654-6964-7/22/\$31.00~\copyright 2022 IEEE \hfill} \hspace{\columnsep}\makebox[\columnwidth]{ }}
\maketitle
\IEEEpubidadjcol

\begin{abstract}
 Image enhancement algorithms are very useful for real world computer vision tasks where image resolution is often physically limited by the sensor size. 
 While state-of-the-art deep neural networks show impressive results for image enhancement, they often struggle to enhance real-world images. 
 In this work, we tackle
 a real-world setting: inpainting of images from Dunhuang caves. %
 The Dunhuang dataset consists of murals, half of which suffer from corrosion and aging. These murals feature a range of rich content, such as Buddha statues, bodhisattvas, sponsors, architecture, dance, music, and decorative patterns designed by different artists spanning ten centuries, which makes manual restoration challenging.
 We modify %
 two different existing methods (CAR, HINet) that are based upon state-of-the-art (SOTA) super resolution and deblurring networks. 
 We show that those can successfully inpaint and enhance these deteriorated cave paintings. 
 We further show that a novel combination of CAR and HINet, resulting in our proposed inpainting network (ARIN), is
 very robust to external noise, especially Gaussian noise. 
 To this end, we present a quantitative and qualitative comparison of our proposed approach with existing SOTA networks and winners of the Dunhuang challenge. 
 One of the proposed methods (HINet) represents the new state of the art and outperforms the \nth{1} place of the Dunhuang Challenge, while our combination ARIN, which is robust to noise, is comparable to the \nth{1} place. We also present and discuss qualitative results showing the impact of our method for inpainting on Dunhuang cave images. 

\end{abstract}


\section{Introduction}
The Mogao Grottoes (also known as Dunhuang cave paintings) \cite{dunhuang} are an internationally recognized cultural heritage site. They represent the caves of the Thousand Buddhas, consisting of 492 temples in the area southeast of the ancient city of Dunhuang in China. These caves contain more than 10000 full frame paintings, which were consecutively created by ancient artists between the \nth{4} and the \nth{14} centuries. The earliest murals date back to over 1600 years, and therefore are of immense heritage value. They provide the opportunity of image studies for historical, artistic, and technological research. 
They are richly abundant in content, serving as a significant site for academic treasure depicting various aspects of medieval politics, economics, culture, arts, religion, ethnic relations, and daily dress in Western China~\cite{unescoref}. The advancement in digitization and annotation technologies have made it possible to explore novel semantic annotation methods to better understand the iconology and iconography of these art works. A very good example is presented in~\cite{wang2018understanding}, where the authors create a framework based on Panofsky's understanding of iconography and iconology studies. Panofsky~\cite{panofsky1939note} stated that Iconology mainly rests on a three-level, art-historical understanding. Based on these three levels, the authors in~\cite{wang2018understanding} divided the annotations of the mural images into ten second-level and twenty nine third level categories in order to understand the correlation between natural description, conventional interpretation and instrinsic meaning.

\begin{figure}[!t]
    \centering
        \includegraphics[width=0.4\textwidth]{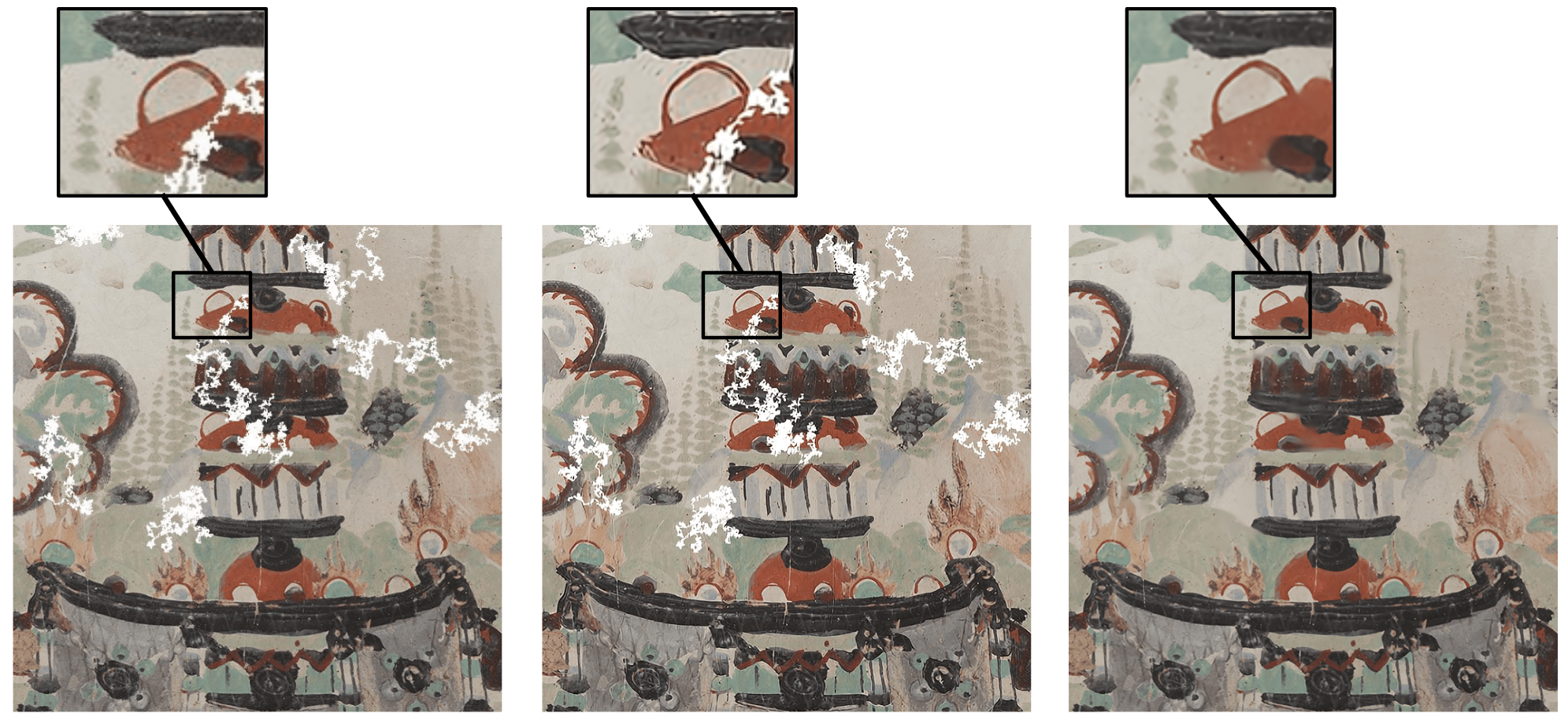}
    \caption{(Left) Damaged mural painting due to aging, (Center) Output when the left image is tested with a \emph{CAR-DF2K} model, (Right) Inpainted image with the proposed method}
    \label{fig:output_overview}
\end{figure}

\begin{figure}[t]
    \centering
    \includegraphics[width=0.24\textwidth, height=0.18\textwidth]{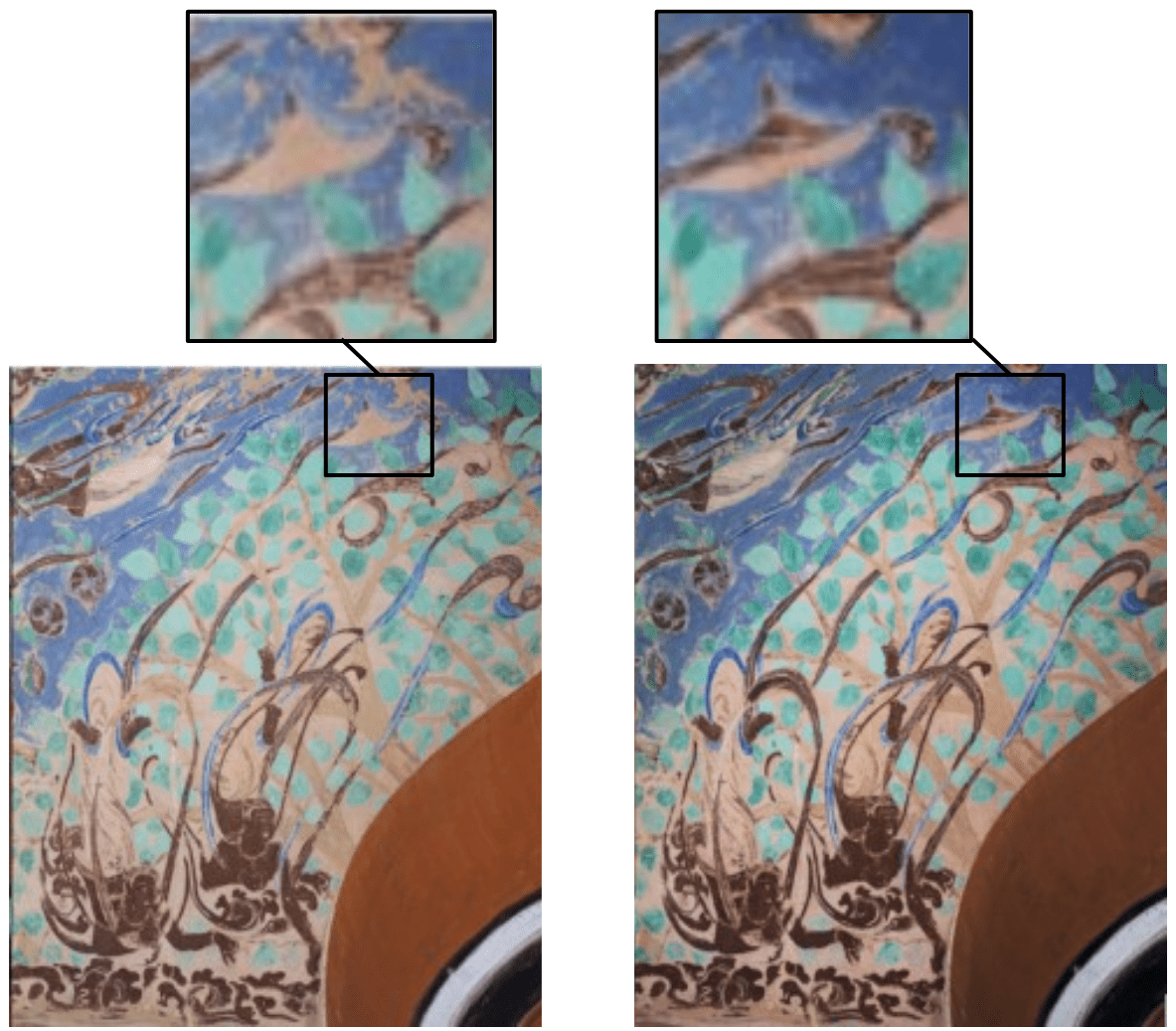}
    \caption{Sample image from Dunhuang cave. Left: Damaged mural painting due to aging, Right: Partial, manual restoration.}
    \label{fig:dunhuangorig}
\end{figure}

\begin{figure}[!t]
    \begin{subfigure}{0.15\textwidth}
        \centering
        \includegraphics[width=\textwidth]{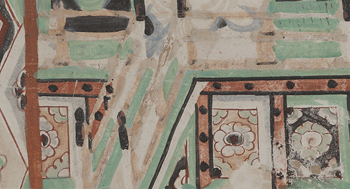}
        \caption{Clean Image}
        \label{fig:clean}
    \end{subfigure}
    \hfill
    \begin{subfigure}{0.15\textwidth}
        \centering
        \includegraphics[width=\textwidth]{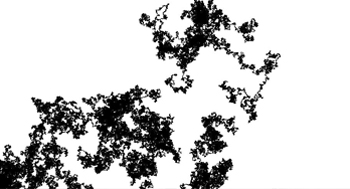}
        \caption{Artificial Mask}
        \label{fig:deterioration}
    \end{subfigure}
    \hfill    
    \begin{subfigure}{0.15\textwidth}
        \centering
        \includegraphics[width=\textwidth]{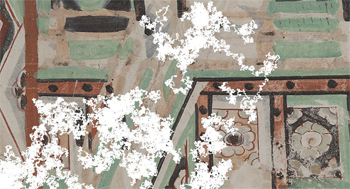}
        \caption{Masked Image}
        \label{fig:deteriorated}
    \end{subfigure}
    
    \caption{An example from Dunhuang Challenge dataset. \subref{fig:clean} shows a clean image; \subref{fig:deterioration} shows a mask that defines one random artifical deterioration; and \subref{fig:deteriorated} shows the artificial deterioration applied on clean image. The challenge considers (\subref{fig:deterioration}, \subref{fig:deteriorated}) for training to recover \subref{fig:clean}}
    \label{fig:datasetfigs}
\end{figure}

The quality of the data highly impacts the image annotation quality. 
These Mogao grottoes have suffered various damages and have degraded over thousands of years. The task of filling this lost information in the images is called image inpainting, an example is shown in \cref{fig:output_overview}. Various researchers in this field so far have done this manually, which is a highly challenging and time-consuming task, example of which can be observed in \cref{fig:dunhuangorig}. This task can be also done automatically using computer vision techniques. To instigate possible image restoration solutions for further annotation and better understanding of iconology, Dunhuang academy organized the first Dunhuang Challenge~\cite{dunhuangpaper} with 600 images of artworks. One sample from the challenge can be observed in \cref{fig:datasetfigs}, where pairs of \cref{fig:deterioration} and \cref{fig:deteriorated} are used for training the algorithm and then tested on unseen deteriorated images. 

Image inpainting algorithms use image deterioration as prior information, while blind image inpainting does not use any prior deterioration information and only uses the deteriorated image. These algorithms work on images from real world taken in a controlled environment because they were exclusively trained on such images. When tested on artwork images (like the Mogao grottoes), they perform below average and do not generalize across domains. 
One such example can be seen in \cref{fig:output_overview}, where the CAR~\cite{car_sun2020learned} network that was trained on the DF2K dataset (\emph{CAR-DF2K}) to restore natural images from real world. 
The main task of the Dunhuang challenge is the digital restoration of the missing parts of a mural painting. This problem is an ill-posed because the finer details of the missing parts are historically lost and the true information no longer exists. Therefore, using a clean image (cf.\ \cref{fig:clean}) and its corresponding artificially degraded counterpart (cf.\ \cref{fig:deteriorated}) for modeling can help us come closer to recreating the lost parts. Recent successes have been seen primarily in the area of deep learning-based image enhancement algorithms~\cite{wang2018dunhuang,inpaintjon}. In~\cite{wang2018dunhuang}, the authors propose an image restoration framework that uses Generative Adversarial Networks (GAN) for restoring deteriorated Dunhuang murals. Their proposed GAN network learns the mapping between deteriorated and restored mural textures between the image pairs. Another method that uses GANs~\cite{inpaintjon} train edge- and color-specific generators in an end-to-end manner, following a sequence of learning edges, color palette and color tones in a hierarchical manner. 

In this paper, we adopt two existing techniques: one for image super-resolution and another one for image deblurring. We apply transfer learning to apply those methods to the blind image inpainting task. In transfer learning, models are trained on one data domain and are then fine-tuned on a smaller dataset from another domain. This has the benefit of shorter training times and efficient learning mechanism. Various computer vision algorithms used for classification, detection, pose estimation, etc.\ initialize their model parameters with a model either pretrained on ImageNet~\cite{deng2009imagenet} or COCO~\cite{lin2014microsoft} dataset.  

We also present a combinatorial approach that is more robust to external noise compared to these transfer-learned models. The first network used is Content Adaptive Resampling (CAR)~\cite{car_sun2020learned}. It uses a generative model to inpaint the restored paintings of the Dunhuang grottoes. The CAR network consists of an end-to-end trained image downsampling and upsampling network that is able to recreate missing image parts. The high variability of the input content (along with the styles) of the grottoes motivated us to use a resampling network, where the learned weights adjust the input degradations so that the network can subsequently remove them. The second network is Half Instance Normalization Network (HINet)~\cite{chen2021hinet} which is the first network to integrate instance normalization for image restoration tasks.

In particular, our contributions to approach the challenging task of restoring the damaged mural artworks are: 
(1)~We adopt CAR~\cite{car_sun2020learned} and HINet~\cite{chen2021hinet} for blind inpainting using transfer learning; 
(2)~We propose Adaptive Resampled Instance Normalized (ARIN) Blind Inpainting Network, which performs at par with the state-of-the-art (SOTA) methods while being more robust to Gaussian Noise and JPEG compression artefacts when compared to CAR or HINet.
(3)~We propose new benchmarks based on common image evaluation methods concerning the Dunhuang Grottoes Painting Dataset. With specialized training and super-resolution networks, we inpaint the sub-sampled deteriorated image to achieve the best and second best performance compared to the winner of the challenge.

The remainder of the paper is organized as follows: \cref{sec:meth} describes our proposed network architectures and training procedures. The detailed description of the Dunhuang challenge and dataset is presented in \cref{sec:dunhuangdata}. In \cref{sec:exp}, we discuss our training and evaluation setup along with the evaluation metrics for this work. We present and discuss our results in \cref{sec:results} and \cref{sec:disc}, respectively. We conclude our work in \cref{sec:conclusion}.

\section{Methodology}\label{sec:meth}

\textbf{Content Adaptive Resampling (CAR)}~\cite{car_sun2020learned} consists of three different parts. First, the resampler network that is used to estimate downsampling kernels for each pixel of the input image based on its image contents. Afterwards, the image is downscaled based on the estimated kernels and super-resolved by the super-resolution (SR) network. Due to this training scheme, the images are downscaled in a way that they can be upsampled again with little loss of information. Therefore, the resampling part of CAR learns content adaptive kernels that are useful for SR at the next stage, which motivated us to choose CAR for inpainting. These kernels can be divided into two parts, each serving a specific task. The first kernel estimates the kernel-weights based on the image contents and the second kernel estimates the corresponding pixel offsets for each pixel in the downscaled image. The kernels are learned based on the generated image evaluation of the upscaling network. In order to model the content of the input image, the resampler network is set up with a convolutional layer which is followed by two inverse sub-pixel convolutions (downsample blocks) to downscale the image. Afterwards, a residual network is added where each block consists of two convolutional layers followed by a downsampling block. This is followed by an upsampling block which learns the kernels that are normalized to the one consisting of three convolutional layers followed by a sub-pixel convolution (one upscaling block for each kernel). Each convolutional layer of the resampler network uses Leaky-ReLU as activation function.
The next step of the pipeline is the image downscaling part, where the previously estimated kernels are applied to each pixel of the deteriorated image to obtain the corresponding pixels in the downscaled image. The process of correct positioning of the resampling kernels is done in the exact same way as mentioned in~\cite{car_sun2020learned}. The next block is the upscaling network, which in our case is EDSR network similar to~\cite{car_sun2020learned}.
While the main motivation of the original paper is to obtain a super-resolved image, we reformulate this network architecture for the blind inpainting problem using transfer learning. CAR is trained jointly with the SR network so that it is able to propagate the error between the reconstructed and the ground truth image back to the image resampling network. 

\begin{figure}
    \centering
    \includegraphics[width=0.6\textwidth]{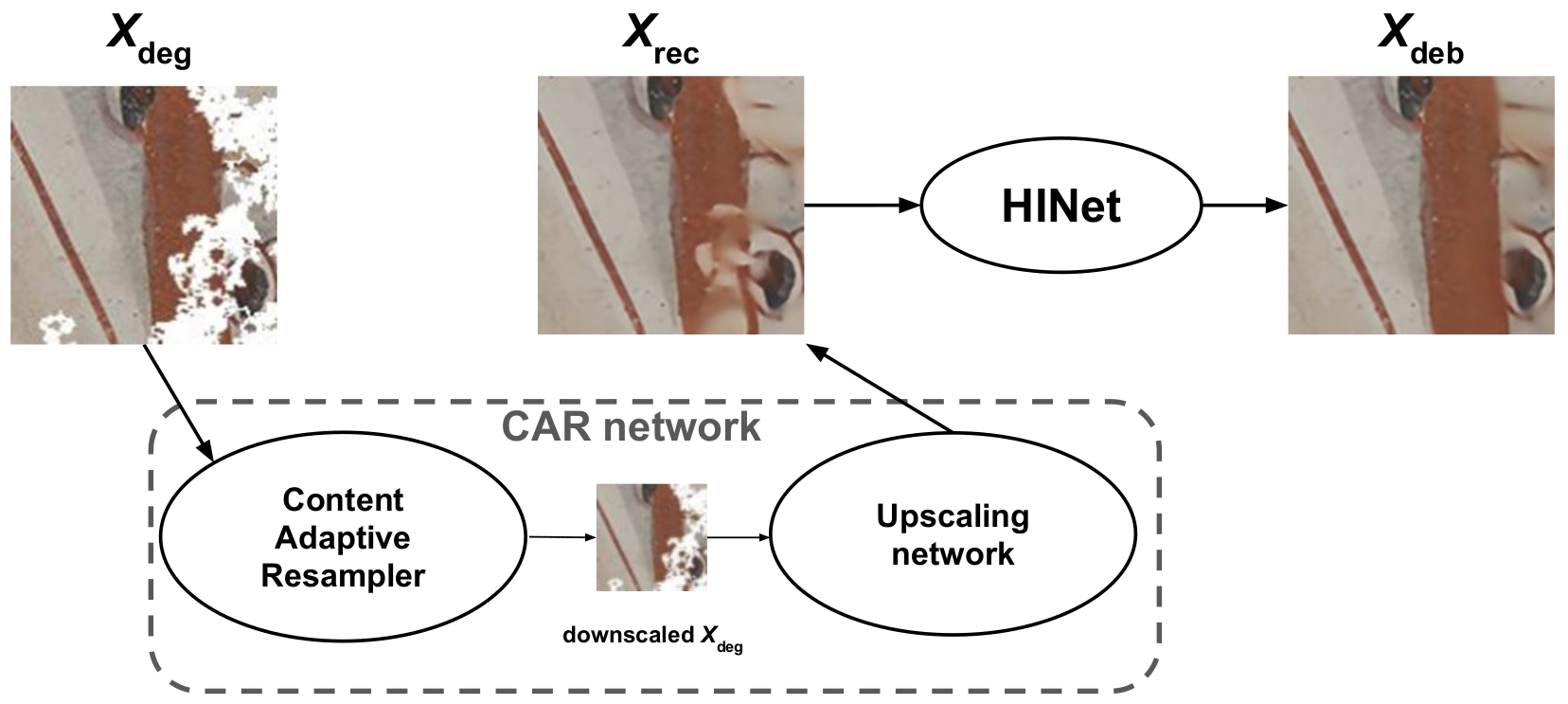}
    \caption{Model Architecture of our proposed ARIN network.}
    \label{fig:car-network}
\end{figure}

\begin{figure}[t]
    \centering
    \begin{subfigure}{0.18\textwidth}
        \centering
        \includegraphics[width=0.95\textwidth]{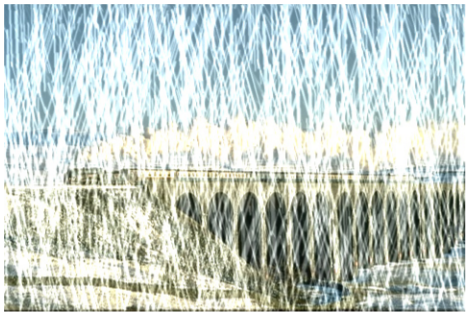}
        \caption{$X_{rain}$}
    \end{subfigure}
    \begin{subfigure}{0.18\textwidth}
        \centering
        \includegraphics[width=0.95\textwidth]{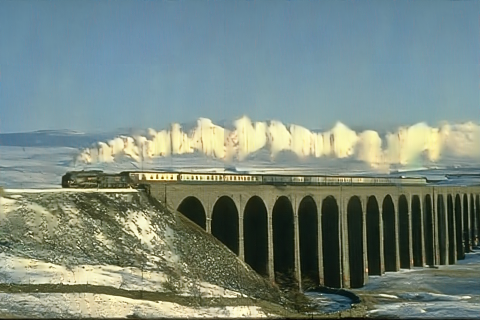}
        \caption{HINet-DR pretrained}
    \end{subfigure}
    
    \caption{An example image of raining as noise and output of pretrained HINet-DR model~\cite{chen2021hinet}.}
    \label{fig:rainingexample}
\end{figure}

\textbf{Half Instance Normalization Network (HINet)}~\cite{chen2021hinet} uses instance normalization instead of batch normalization. Based on their proposed Half Instance Normalization (HIN) blocks, a variational autoencoder network structure with two sequential subnetworks of U-Nets~\cite{ronneberger2015u} is introduced. Similar to a deep-residual network architecture, the HIN Blocks are stacked in each of the encoders to improve the performance. The two U-Nets are then trained in an end-to-end manner. We also adapt this image restoration network for the task of blind inpainting using transfer learning. Based on the networks presented in HINet~\cite{chen2021hinet}, we train two networks: HINet-Derain (HINet-DR) and HINet-Deblur (HINet-DB). The HINet-DR network is trained to remove the raining noise, cf.\ \cref{fig:rainingexample}, while the HINet-DB network is trained for regular deblurring tasks.

\textbf{Adaptive Resampled Instance Normalization (ARIN)}  is motivated from the previous two networks. Due to the kernel learning capability of CAR and image denoising capability of HINet, ARIN learns the weights during the downscaling process to prepare the low-resolution (LR) image inputs for better upsampling and then upsamples it as a denoised SR image. Eventually, this leads towards an end-to-end blind inpainting problem. \cref{fig:car-network} shows the proposed architecture. The degraded mural painting images ($X_\text{deg}$) are downscaled using CAR which helps to improve the image quality by the ARIN network eventually. The learned downsampling part of the CAR network can be applied for factors of two or four. In this work, we use a downscaling factor of two. Furthermore, the size of the downsampling kernel was set to \numproduct{6 x 6}, which was adopted by the original training process consequently. 
A very common issue in image restoration is blurry outputs. In order to mitigate this, we apply the HINet~\cite{chen2021hinet} to deblur and correct missing image information of the reconstructed CAR images. 
 We use pretrained networks for each of the network blocks and then finetune on the specific image inpainting or deblurring task.

\textbf{Loss functions}:
Let $X_\text{rec}$ represent the set of reconstructed image outputs from CAR network. 
$R_\text{deb}^1$ \& $R_\text{deb}^2$ represents the predicted subnetwork (U-Net) outputs of HINet and $X_c$ represents the original clean images. 
To train the image downscaling network in the first step, we use the error between $X_\text{rec}$ and $X_c$. Image perceptual quality is better preserved with mean absolute error (MAE) or $L_1$ norm when compared to mean squared error (MSE)~\cite{hui2018fast}. Furthermore, MAE loss produces better results and can speed up the training process.  Hence, we also adopt the $L_1$ norm loss regarding the blind inpainting problem which can be defined as:
\begin{equation}
 L_\text{DS} = \frac{1}{N} \sum_{p=1}^{N} \lvert X_{\text{rec}}^p - X_c^p\rvert \;, 
\end{equation}
where $N$ is the number of corresponding patches from $X_c$ and $X_\text{rec}$ respectively. 

For the second step, the reconstructed image is deblurred by HINet, which minimizes the error between the reconstructed deblurred image and the corresponding ground truth image. The HINet is trained with the following loss function:
\begin{equation}
    L_\text{HINet} = - \sum_{i = 1}^{2} PSNR((R_{\text{deb}}^i + X_{\text{rec}}^i), X_c)  \;,
\end{equation}
where i represents each U-Net of HINet.

\section{Dunhuang Challenge Dataset}\label{sec:dunhuangdata}
The Dunhuang Challenge~\cite{dunhuangpaper} was organized as part of the ICCV 2019 workshop on E-heritage. The dataset consists of 600 images from the mural paintings on Cave No.\ 7 as it has a mixture of deteriorated and preserved regions. Each image has a focus on various themes, such as Buddha, architecture, decoration and human. 500 images were used for training and 100 images for testing. The challenge organizers generated artificial deterioration by creating randomized masks (cf.\ \cref{fig:deterioration}) and also motivated the participants to create their own deteriorations. All the participants were provided with 500 triplets (cf.\ \cref{fig:datasetfigs}), where each triplet comprised of ground truth (clean image), randomly generated mask (deterioration) and input image (deteriorated image) for training and validation purposes. It is important to note that there were no ground truths provided for the test set for the challenge. However, we were able to obtain the ground-truth for the test-sets after contacting the organizers, and therefore our evaluations are done on the original test-set.

\section{Experimental Setup} \label{sec:exp}
\subsection{Training} \label{sec:exp_train}
The networks for this work rely on their pretrained models (HINet ~\cite{chen2021hinet}, CAR ~\cite{car_sun2020learned}) and were trained further for a different number of iterations (cf.\ \cref{tab:iterations}), using the loss functions as described above. For both the networks, we use the ADAM optimizer with an initial learning rate of $10^{-4}$ and $\beta_1\ = 0.9$, $\beta_2 = 0.999$.

\textbf{CAR} network was trained with a batch size of 10. We cropped the train and validation images in blocks of \numproduct{128 x 128} with minimal overlap to take advantage of the respective full image resolution. 19880 patches were used for training and 3976 for validation. The model was implemented in PyTorch and then trained on a single CUDA-enabled GPU with a memory capacity of 11 GB using Google Colab.

\textbf{HINet} was trained with a batch size of 4. However, for HINet training, to retain the input size at the output, all 500 training images were first resized to make it compatible for CAR input (divisible factor of 6 x 6). All these images were then split into 128 x 128 patches resulting in 12009 patches. We use 10\% of the total patches for validation and remaining were used for training. The model was implemented in PyTorch and then trained on a single CUDA-enabled GPU with a memory capacity of 6 GB on Windows 10.

\subsection{Evaluation Metrics}\label{sec:exp_evaluation}
In total, we train four networks: CAR, HINet-DR, HINet-DB, and ARIN. The baseline for our work is \emph{CAR-DF2K}, a CAR~\cite{car_sun2020learned} network trained on the DF2K dataset - a merged dataset from DIV2K~\cite{Ignatov_2018_ECCV_Workshops} and Flickr2K~\cite{Lim_2017_CVPR_Workshops}. The performance of all our networks is compared with the top three winners of the Dunhuang Challenge~\cite{dunhuangpaper}. The evaluation metrics used to measure the quality of the reconstructed images are Peak-Signal-to-Noise-Ratio (PSNR) and Structural Similarity (SSIM). We also present the evaluation with another metric called Learned Perceptual Image Parch Similarity (LPIPS)~\cite{zhang2018unreasonable} -- it has been shown to be closest to human perception, and therefore can be established as a more relevant metric. Dissimilarity SSIM (DSSIM) is used to compare our results with the winner of the Dunhuang image inpainting challenge. According to the paper of the challenge DSSIM was calculated as:

\begin{equation}
    \text{DSSIM}_\text{challenge}(I, \hat{I}) = 1 - \text{SSIM}(I, \hat{I}) 
\end{equation}
with $I$ and $\hat{I}$ are the ground truth and the reconstructed image respectively.
To retain the input size at the output, all the test images were resized to make them compatible for CAR input (divisible factor of 6 x 6) and given in their adjusted full-size to the networks since they were applied in a fully-convolutional manner.

\begin{table}[!t]
\centering
\caption[Dunhuang results]{Comparison of the best proposed method with the Dunhuang Challenge~\cite{dunhuangpaper} winners. `--' indicates that the values are not available.}      
\label{tab:results}
\begin{tabular}{@{}lcccc@{}}
\toprule
    Network & PSNR\(\uparrow\) & SSIM\(\uparrow\) & DSSIM \(\downarrow\) & LPIPS \(\downarrow\) \\ \midrule
    CAR-DF2K & 16.65 & 0.67 & 0.33 & 0.33 \\
    VIDAR (\nth{1} Rank) & -- & 0.87 & 0.13 & -- \\
    tomcarrot (\nth{2} Rank) & -- & 0.84 & 0.16 & -- \\
    UIUC-AE (\nth{3} Rank) & -- & 0.84 & 0.16 & -- \\
    CAR & 27.70 & 0.84 & 0.16 & 0.17 \\
    HINet-DR & 24.40 & 0.83 & 0.17 & 0.16 \\
    \textbf{HINet-DB} & \textbf{33.32} & \textbf{0.90} & \textbf{0.10} & \textbf{0.08} \\
    ARIN & 31.00 & 0.86 & 0.14 & 0.15 \\
\bottomrule
\end{tabular}
\end{table}

\begin{table}[!t]
\centering
\caption[Dunhuang results]{Effect of training iterations for all the proposed architectures. Multiple values in the `Iterations' column indicate the number of iterations used for training two networks separately as mentioned in \cref{sec:exp_train}.}      
\label{tab:iterations}
\begin{tabular}{@{}lcccc@{}}
\toprule
    Network & Iterations & PSNR\(\uparrow\) & SSIM\(\uparrow\) & LPIPS \(\downarrow\) \\ \midrule
    CAR & 4k & 24.40 & 0.84 & 0.17 \\
    CAR & 6k & 27.40 & 0.84 & 0.17 \\
    CAR & 8k & 27.00 & 0.84 & 0.17 \\
    ARIN & 8k, 50k & 30.90 & 0.86 & 0.15 \\
    ARIN & 8k, 100k & 30.80 & 0.86 & 0.15 \\
    ARIN & 6k, 50k & 30.90 & 0.86 & 0.15 \\
    ARIN & 6k, 100k & 31.00 & 0.86 & 0.15 \\
\bottomrule
\end{tabular}
\end{table}

\section{Quantitative Evaluation}\label{sec:results}
\subsection{Comparison with state of the art}\label{subsec:compsota}
\Cref{tab:results} compares our results with SOTA. HINet-DR performs slightly better than the basic CAR approach in terms of human perception metric (LPIPS). However, in terms of PSNR values, CAR outperforms HINet-DR. One possible explanation could be that there are some color differences between pixels in the picture, which may not be as relevant for human perception. Furthermore, the SSIM is a bit worse which underlines the previous assumption. According to DSSIM (the metric used in the challenge), proposed ARIN and HINet-DB outperform the \nth{2} and \nth{3} ranks of the challenge. In addition, HINet-DB is superior to all other methods in all metrics and thus represents the new state of the art for the 2019 Dunhuang challenge. Since we did not have the LPIPS metric recorded for the challenge winners, we could not compare the same. 

\begin{figure*}
    \centering
    \begin{subfigure}{0.13\textwidth}
        \centering
        \includegraphics[width=0.85\textwidth]{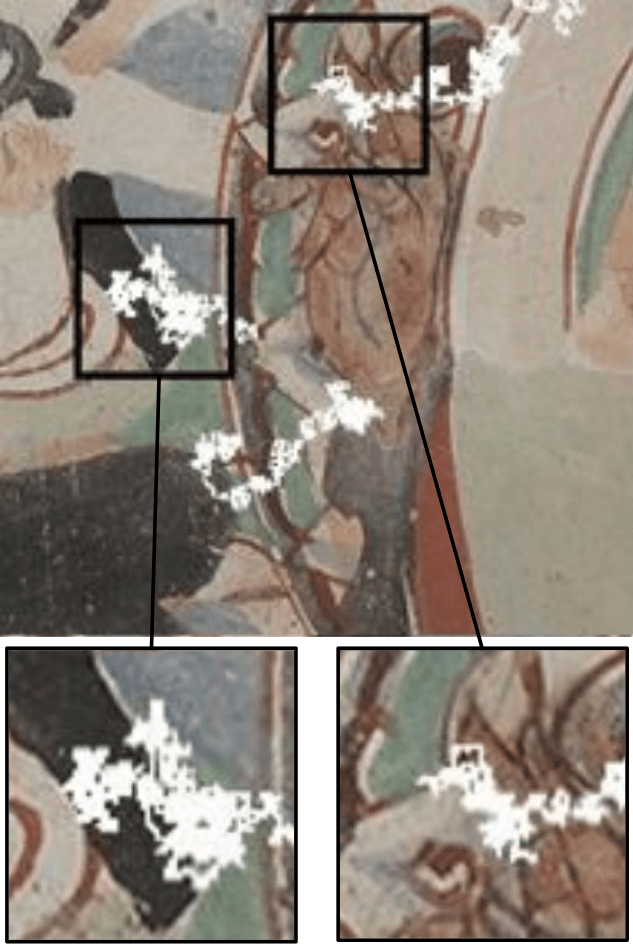}
        \caption{$X_{deg}$}
    \end{subfigure}
    \begin{subfigure}{0.13\textwidth}
        \centering
        \includegraphics[width=0.85\textwidth]{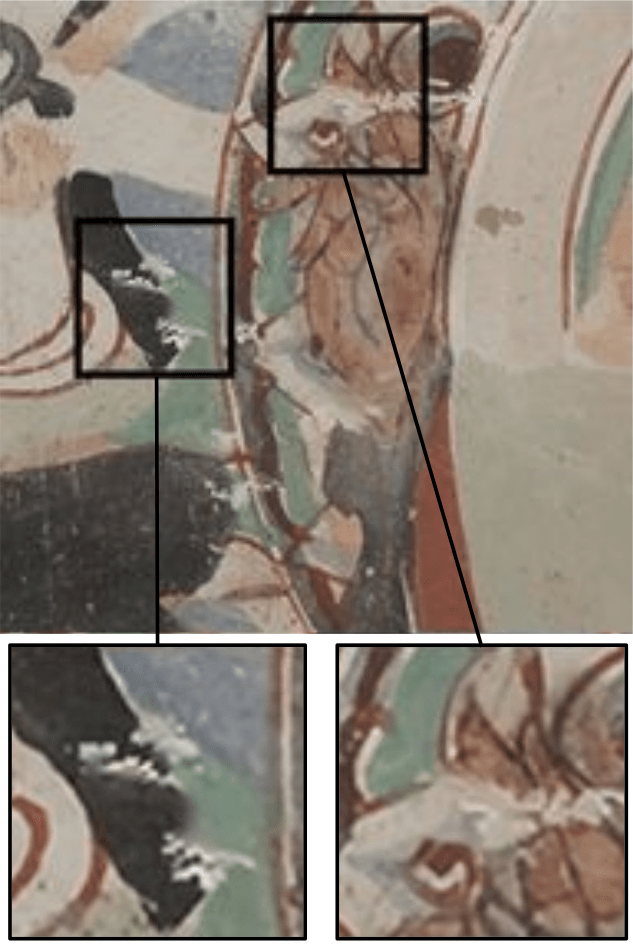}
        \caption{HINet-DR}
    \end{subfigure}
     \begin{subfigure}{0.13\textwidth}
        \centering
        \includegraphics[width=0.85\textwidth]{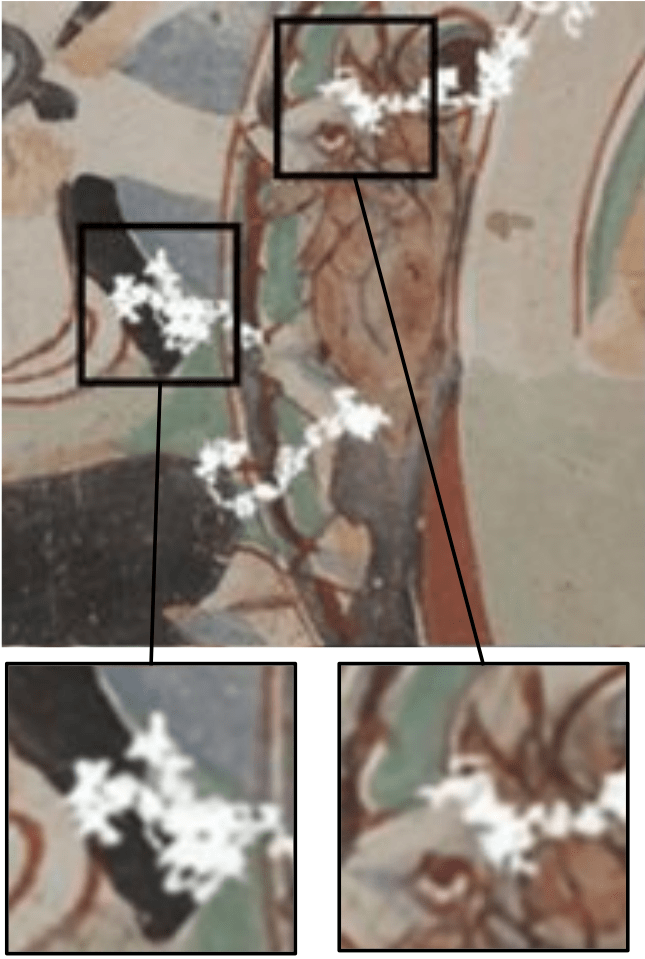}
        \caption{CAR-DF2K}
    \end{subfigure}
    \begin{subfigure}{0.13\textwidth}
        \centering
        \includegraphics[width=0.82\textwidth]{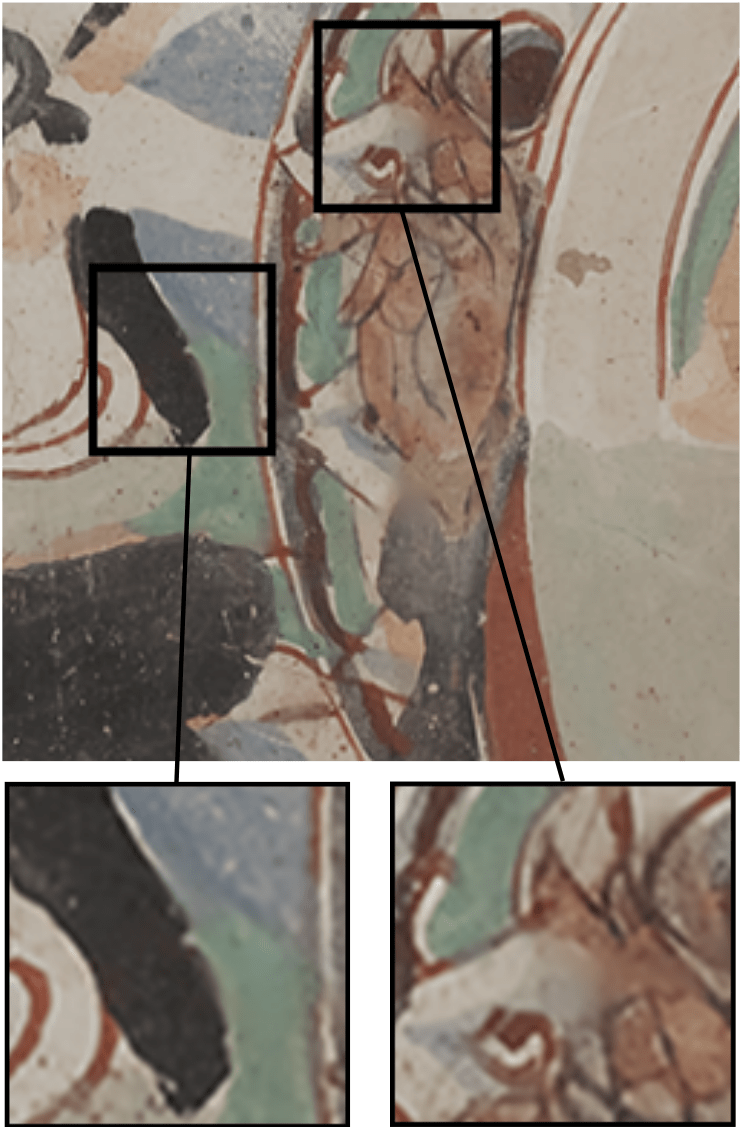}
        \caption{HINet-DB}
    \end{subfigure}
    \begin{subfigure}{0.13\textwidth}
        \centering
        \includegraphics[width=0.85\textwidth]{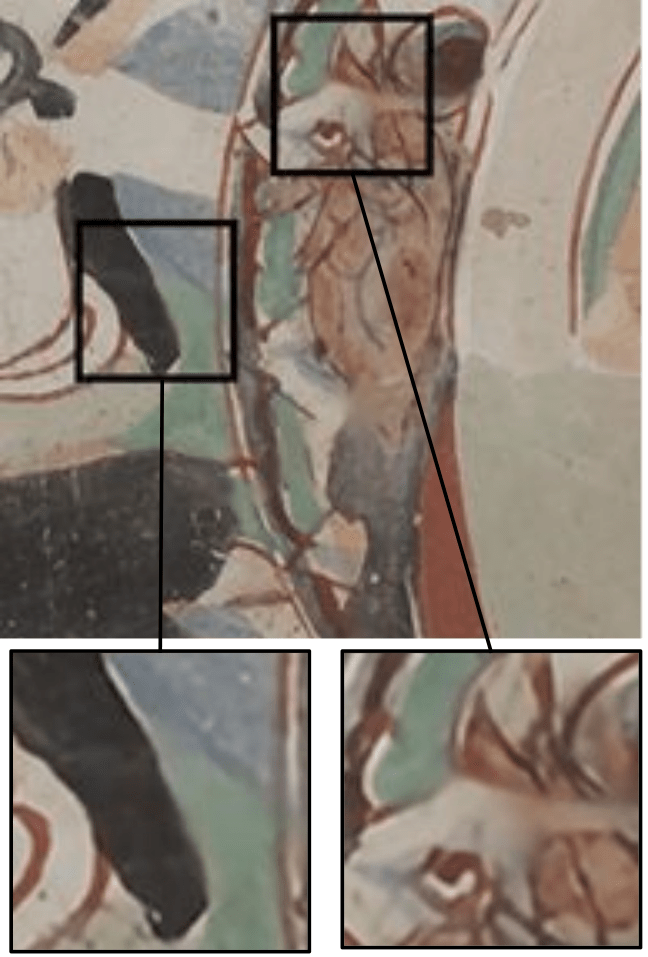}
        \caption{CAR}
    \end{subfigure}
    \begin{subfigure}{0.13\textwidth}
        \centering
        \includegraphics[width=0.85\textwidth]{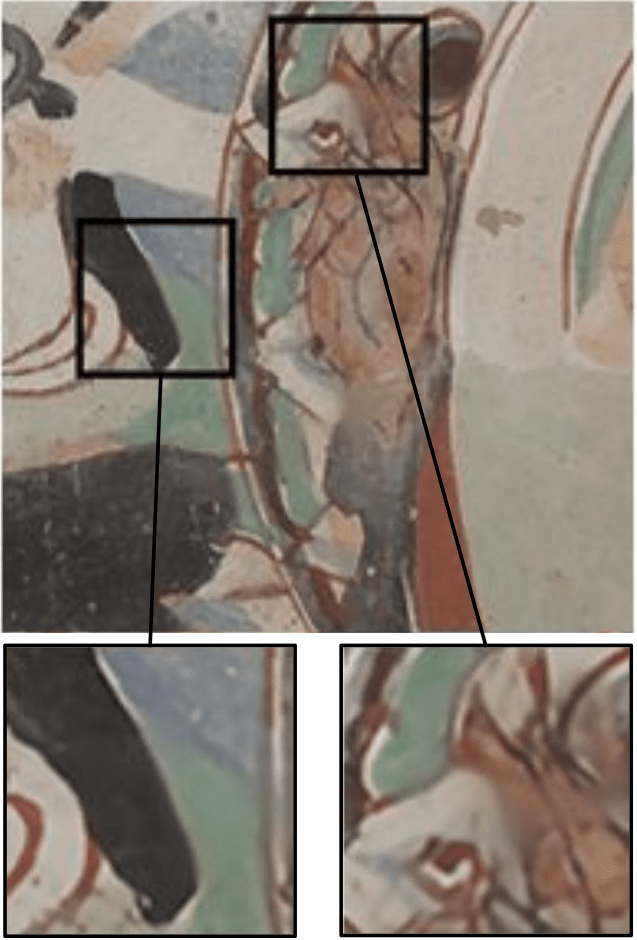}
        \caption{ARIN}
    \end{subfigure}
    \begin{subfigure}{0.13\textwidth}
        \centering
        \includegraphics[width=0.85\textwidth]{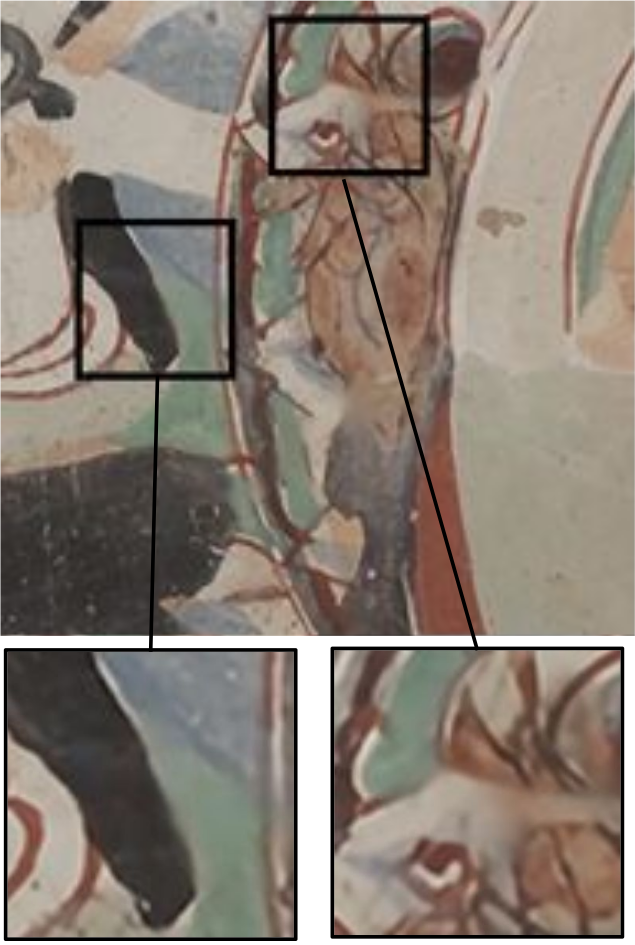}
        \caption{GT}
    \end{subfigure}
    \caption{Qualitative evaluation of proposed method. ARIN is the closest to the GT. The regions in black are zoomed for better visualization.}
    \label{fig:qualeval}
\end{figure*}

\subsection{Effect of training iterations}\label{subsec:efftrain}
We study the effect of the number of training iterations for our networks.   \Cref{tab:iterations} shows that overall there is not a significant difference. The CAR network trained for 6k iterations performs best in each evaluation metric. At 8k iterations, the network does not improve anymore but rather worsens the output. In combination with the deblurring network, a big improvement can be observed for our proposed network (ARIN). Generally, ARIN performs better than CAR. For ARIN, HINet is trained on top of CAR outputs. CAR outputs obtained by running 6k iterations are again slightly better than the ones trained on CAR outputs of 8k iterations. Furthermore, while there is a lot of improvement from CAR to ARIN, there is a low noticeable difference in the evaluation between ARIN when HINet is trained for 50k or 100k. Based on these observations of training iterations and performance, ARIN (6k, 100k) model is the best model. 

\subsection{Effect of External Noise}\label{subsec:noise}
To test the robustness of our proposed method on additional real-world degradations such as noise resulting from low-light conditions or compression artifacts, we added Gaussian noise and JPEG compression to the test data. We consider two cases for Gaussian noise: low and extreme. For the low case, we use a standard deviation of 8 and for the extreme case, standard deviation of 16. \Cref{tab:rw_degradation} shows that in the low case, our proposed method ARIN is robust, while the performance deteriorates a little in the extreme case.

For JPEG compression, ARIN provides decent results, however some image information is lost due to the compression and the LPIPS metric is in favor of HINet-DB while ARIN is superior in terms of PSNR and SSIM. The reconstructed images themselves do not show the typical checker pattern artifacts of JPEG compression, thus ARIN successfully cleans the images. Furthermore, there is only a small difference between the reconstructed outputs when using different JPEG quality index of 60 or 30.

\section{Qualitative Results}\label{sec:disc}
\subsection{Comparison with SOTA}\label{sec:qualsota}
As shown in \cref{fig:qualeval}, the baseline approach does not restore the clean image. This is likely due to the domain shift in the Dunhuang paintings. The HINet-DR model partially fails at inpainting, however is able to fill few parts of the lost image information. In contrast, all our proposed methods (CAR, ARIN, HINet-DB) successfully inpaint the missing information and enhances the degraded image.

\subsection{Effect of External Noise}\label{sec:qualnoise}
\begin{figure}
    \centering
    \begin{subfigure}{0.11\textwidth}
        \centering
        \includegraphics[width=0.8\textwidth]{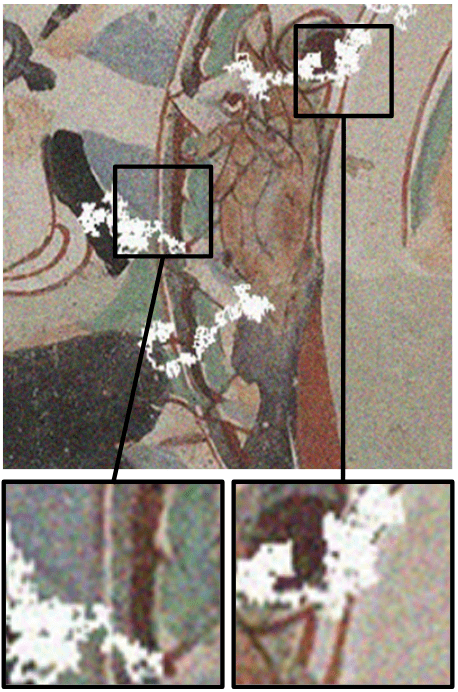}
        \caption{GN(16)}
        \label{fig:gnnoisequaleval_a}
    \end{subfigure}
     \begin{subfigure}{0.11\textwidth}
        \centering
        \includegraphics[width=0.8\textwidth]{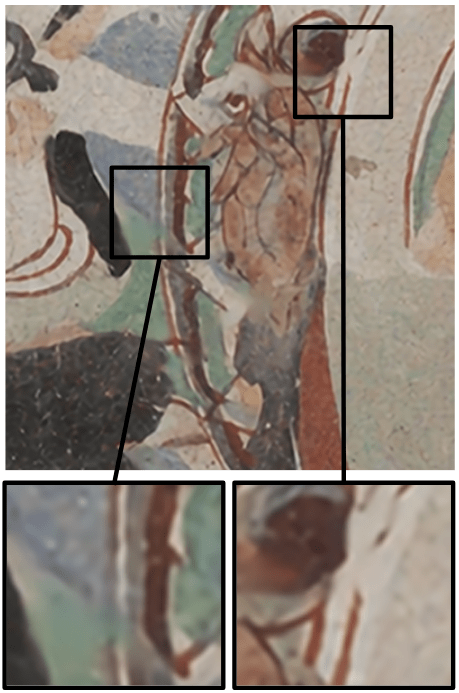}
        \caption{ARIN}
        \label{fig:gnnoisequaleval_b}
    \end{subfigure}
    \begin{subfigure}{0.11\textwidth}
        \centering
        \includegraphics[width=0.8\textwidth]{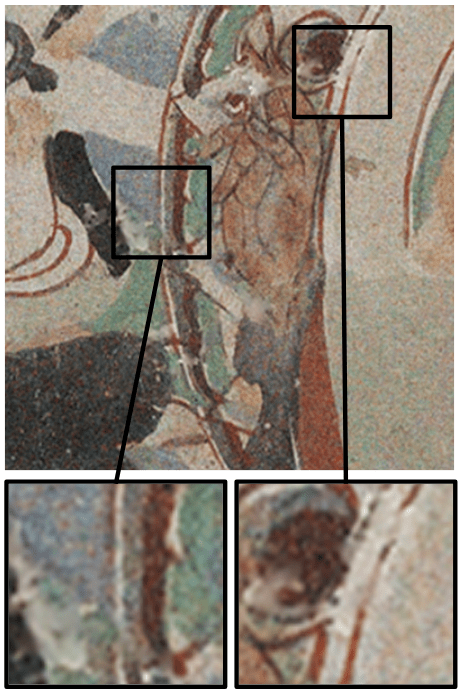}
        \caption{HINet-DB}
        \label{fig:gnnoisequaleval_c}
    \end{subfigure}
    \begin{subfigure}{0.11\textwidth}
        \centering
        \includegraphics[width=0.8\textwidth]{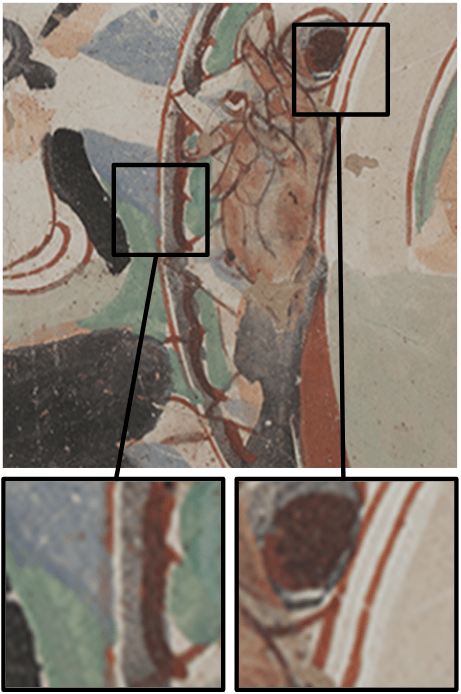}
        \caption{GT}
        \label{fig:gnnoisequaleval_d}
    \end{subfigure}
    \caption{Qualitative impact of Gaussian noise on the methods. ARIN is mainly robust towards the extreme case of Gaussian noise as compared to HINet which is the best performing method. The regions in black are zoomed for better visualization.}
    \label{fig:gnnoisequaleval}
\end{figure}

As described in \cref{subsec:noise}, for PSNR and SSIM, our proposed ARIN network quantitatively outperforms all the networks (including the best performing HINet-DB) for both Gaussian Noise and JPEG noise in low and extreme cases. Comparing the LPIPS metric, it performs at par or better than HINet-DB for Gaussian Noise, however does not beat HINet-DB for JPEG compression noise.
\cref{fig:gnnoisequaleval_a} shows one example image from the test dataset added with large Gaussian noise and \cref{fig:gnnoisequaleval_d} shows the corresponding ground truth image. \Cref{fig:gnnoisequaleval_b,fig:gnnoisequaleval_c} show that the output of HINet-DB has green and red dots due to the Gaussian noise while ARIN successfully removes the Gaussian noise. However, this comes at the cost of a possible over-smoothening of the image content.

\begin{figure}
    \centering
    \begin{subfigure}{0.11\textwidth}
        \centering
        \includegraphics[width=0.8\textwidth]{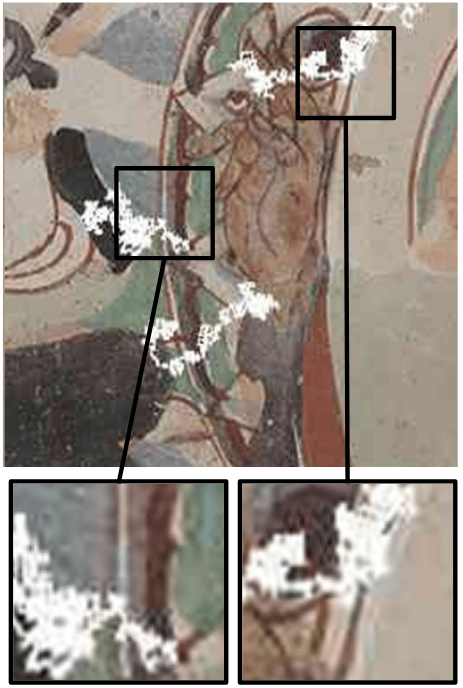}
        \caption{JPEG(30)}
        \label{fig:jpegnoisequaleval_a}
    \end{subfigure}
     \begin{subfigure}{0.11\textwidth}
        \centering
        \includegraphics[width=0.8\textwidth]{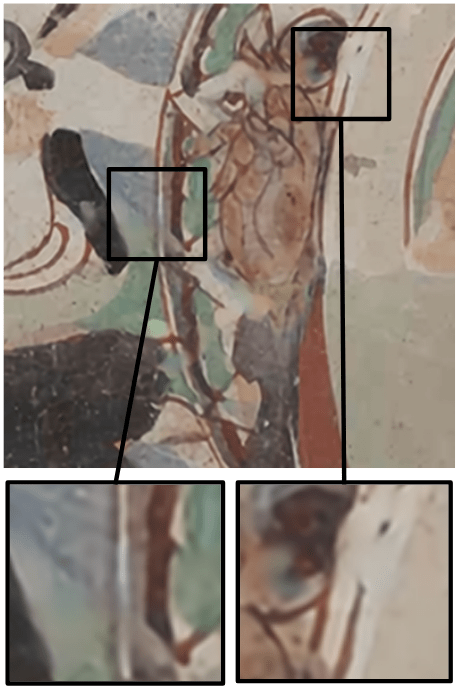}
        \caption{ARIN}
        \label{fig:jpegnoisequaleval_b}
    \end{subfigure}
    \begin{subfigure}{0.11\textwidth}
        \centering
        \includegraphics[width=0.8\textwidth]{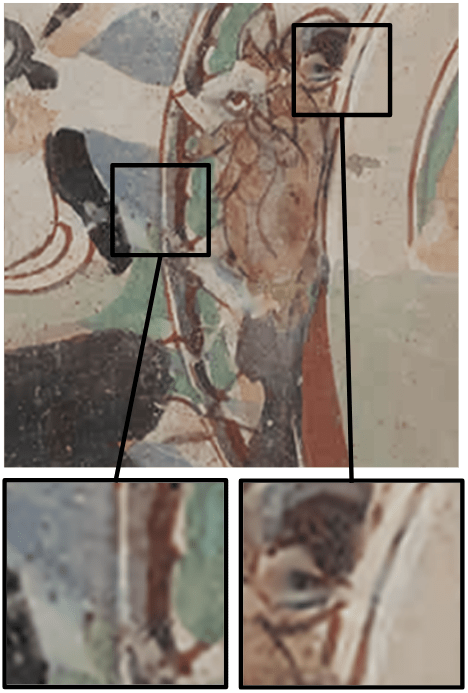}
        \caption{HINet-DB}
        \label{fig:jpegnoisequaleval_c}
    \end{subfigure}
    \begin{subfigure}{0.11\textwidth}
        \centering
        \includegraphics[width=0.8\textwidth]{figures/xnoisegroundtruth.png}
        \caption{GT}
        \label{fig:jpegnoisequaleval_d}
    \end{subfigure}
    \caption{Qualitative impact of JPEG compression noise on the methods. ARIN is mainly robust towards the extreme case of JPEG compression noise as compared to HINet which is the best performing method. The regions in black are zoomed for better visualization.}
    \label{fig:jpegnoisequaleval}
\end{figure}

\cref{fig:jpegnoisequaleval_a} shows one example image from the test dataset applied with high JPEG compression and \cref{fig:jpegnoisequaleval_d} shows the corresponding ground truth image. \cref{fig:jpegnoisequaleval_b,fig:jpegnoisequaleval_c} show that the output of HINet-DB is visually better in the terms of sharpness and preservation of the small details while the output of the ARIN network seems over-smoothed, validating the quantitative numbers from \cref{tab:rw_degradation} for the LPIPS metric. 

\begin{table}[!t]
\centering
\caption[Real-world degradation]{Results of HINet-DB (50k) (HDB) and ARIN-SEP (6k, 100k) on simulated real-world degradations represented by Gaussian noise (GN) with a standard deviation of 8 and 16, and JPEG compression with a quality level of 30 and 60.  '--' represents baseline without any noise.}
\label{tab:rw_degradation}
\begin{tabular}{@{}lcc|cc|cc@{}}
\toprule
Noise & \multicolumn{2}{c}{PSNR\(\uparrow\) } & \multicolumn{2}{c}{SSIM\(\uparrow\)} & \multicolumn{2}{c}{LPIPS\(\downarrow\)} \\ \midrule
 & ARIN & HDB & ARIN & HDB &  ARIN  & HDB\\
\midrule
-- & 31.00 & \textbf{33.32} & 0.86 & \textbf{0.90} & 0.15 & \textbf{0.08} \\
GN(8) & \textbf{30.40} & 28.80 & \textbf{0.84} & 0.78 & \textbf{0.15} & \textbf{0.15} \\
GN(16) & \textbf{28.50} & 23.20 & \textbf{0.76} & 0.55 & \textbf{0.21} & 0.35 \\
JPEG(60) & \textbf{29.90} & 28.20 & 0.84 & 0.84 & 0.18 & \textbf{0.13} \\
JPEG(30) & \textbf{28.30} & 25.30 & \textbf{0.80} & 0.77 & 0.23 & \textbf{0.21} \\ \bottomrule
\end{tabular}
\end{table}

\section{Conclusion}\label{sec:conclusion}
In this paper, we propose multiple approaches for the task of blind image inpainting in the context of restoring degraded Dunhuang Grottoes paintings. CAR, an integral part of our proposed method ARIN, helps to learn the downscaling due to which the images are downscaled in a way that they can be upsampled again with the best possible result and lower loss of image information. Although there is a small loss of details in the reconstructed images due to the network characteristics and the image downscaling, it can be observed that lost artifacts are reconstructed extremely well even in large areas, which is reflected by the evaluation metrics. Furthermore, our method HINet-DB beats the state-of-the-art method in all evaluated metrics. Our proposed method, ARIN, is shown to be robust to simulated real-world degradations such as Gaussian noise and JPEG compression in low as well as extreme cases. As one of our future works, we plan to train the ARIN network in an end-to-end fashion. 

\section{Acknowledgment} \label{sec:acknowledge}
We thank the authors of ``Dunhuang Grottoes Painting Dataset and Benchmark''~\cite{dunhuangpaper} for sharing their train and test datasets, which enabled us to create a valid evaluation for our proposed method. The paper is partially funded by Odeuropa EU H2020 project under grant agreement No. 101004469

\bibliographystyle{IEEEtran}
\bibliography{IEEEabrv,biblio}

\end{document}